\pdfoutput=1

\RequirePackage{amsmath}
\documentclass{llncs}
\usepackage{graphicx}
\usepackage{multirow}
\usepackage{tabularx,ragged2e}
\usepackage{makecell}
\usepackage[labelformat=simple]{subcaption}
\usepackage{booktabs}
\usepackage{amsfonts,amssymb} 
\usepackage[ruled,linesnumbered]{algorithm2e}
\usepackage{float}

\newcolumntype{C}{>{\Centering}X} 

\usepackage{booktabs} 

\newcommand{\RN}[1]{%
  \textup{\expandafter{\romannumeral#1}}%
}

\begin{document}

\title{A Robust Model for Trust Evaluation during Interactions between Agents in a Sociable Environment}

\author{Qin Liang\inst{1,2}, Minjie Zhang\inst{1}, Fenghui Ren\inst{1} \and Takayuki Ito\inst{2}}

\institute{University of Wollongong, Wollongong \inst{1}, Australia  \\ Nagoya Institute of Technology \inst{2}, Japan  \\
\email{ql948@uowmail.edu.au, \{minjie, fren\}@uow.edu.au}\inst{1}  \\ \email{ito.takayuki@nitech.ac.jp} \inst{2} 
}

\maketitle

\begin{abstract}

Trust evaluation is an important topic in both research and applications in sociable environments. This paper presents a model for trust evaluation between agents by the combination of direct trust, indirect trust through neighbouring links and the reputation of an agent in the environment (i.e. social network) to provide the robust evaluation. Our approach is typology independent from social network structures and in a decentralized manner without a central controller, so it can be used in broad domains. 

\end{abstract}

\section{Introduction}

In general, trust is a subjective belief that an entity (evaluator) evaluates the reliability of another entity's (evaluatee's) future behaviours in a sociable environment, based on past interactions between them, and available knowledge of the evaluatee in the surrounding environment. Nowadays, trust evaluation has become an important research topic and is widely used in a variety of sociable scenarios, such as in social networks \cite{hang2009propagate} \cite{liu2014assessment}, e-commerce \cite{wang2015trust} \cite{fang2015multi}, P2P networks \cite{zhao2013aggregation}, web-services \cite{wang2017trust}, internet of things \cite{nitti2014management}, and crowdsourcing \cite{Yu2013reputation}, and it is also an important basis of selecting suitable interaction partners for trustors to maximize its benefits. 

Many approaches and models have been proposed and developed for trust evaluation from different perspectives, such as the CertProp approach proposed by Hang and et al \cite{hang2009propagate}, which considered concatenation, aggregation, and selection operators to solve the trust propagation problem in social networks; the SWORD approach proposed by Yu and et al \cite{Yu2013reputation}, which used reputation-aware decision-making to solve task assignment problem in crowdsourcing platforms; the (dis)trust framework proposed by Fang and et al \cite{fang2015multi}, which considered interpersonal and impersonal aspects of trust and distrust to model user preference in recommendation systems; and the algorithm proposed by Wang and et al \cite{wang2017trust}, which combined both user preference and service trust to solve service composition problems in web service.  

A sociable environment is a decentralized and open environment. Each member has only a local view without a centralized controller. In such an environment, how to design a trust model by the consideration of dynamic and decentralized features with only local views of members for providing robust and automatic trust evaluation is a challenging question. The motivation of this paper is to attempt to provide a solution to this question. 

In this research, a $trustor$ and a $trustee$ are used to represent an evaluator and an evaluatee, respectively, and all participators of the environment are modeled as intelligent agents. This paper proposes a robust model for trust evaluation in a sociable environment. The model includes three modules, which are (1) a direct trust evaluation module based on agents' interaction experiences, (2) an indirect trust evaluation module through neighbours, and (3) a reputation collection module.

The merits of our model have twofold. \textbf{(1)} In our model, a trustor evaluates a trustee for a requested task not only from its direct experiences with the trustee, but also considering the experiences from neighbours with the trustee for the same task so as to avoid subjective evaluation. \textbf{(2)} Our model not only considers both direct and indirect trusts on a trustee for the requested task, but also takes the consideration of the reputation value of the trustee (given by other members who have interactions with the trustee for other tasks). By utilizing the reputation value to our trust evaluation, the evaluator can get an indication about the comprehensive performance of the trustee so as to enhance the robustness of the evaluation result. 

The rest of this paper is organised as follows. Section 2 introduces the problem and gives the formal definitions. Section 3 describes the principle of our model and gives the detail design of three trust evaluation modules. Section 4 is related work. Section 5 concludes the paper and outlines the future work.

\section{Problem Description and Definitions}
\subsection{Problem Description}

In this paper, a Multi-Agent System (MAS) is used to represent the sociable environment, where each entity is represented as an $agent$, which can be a human agent, or an intelligent agent, and these agents can communicate with each other to accomplish some kind of tasks. The communication between agents is represented as an $interaction$. To accomplish the task, an agent needs to evaluate the trustworthiness of others through interactions. 

An agent will first calculates the direct trust based on past interactions with potential partners. If there is no direct interaction or only has few interactions, the agent needs to calculate the indirect trust based on the recommendations from trusted agents. Besides, the agent can also refer to the reputation of a potential partner given by other members. For example, in an e-commerce environment, a buyer will consider previous transaction experiences, other buyers' ratings and rating from the e-commerce platform before making a deal with a seller. Therefore, trust evaluation is a crucial foundation in interactions among agents in a sociable environment.

In this paper, a sociable environment is modeled as a weighted directed graph, shown in Fig. 1. A node represents an agent with some task interactions. The directed edge represents a direct trust relationship between a $trustor$ and a $trustee$ based on the past interactions between them. The trust relationship between two agents can be an one-way or a two-way relationship based on specific sociable scenes. For example, in an e-commerce environment, if a buyer and a seller both rate the transactions, then the trust relationship between them is a two-way relationship.

Fig. 1 shows a simple example of trust evaluation. Suppose at some point of time, agent $A_{1}$ proposes a request to seek for a task partner for task category $TK_{3}$. Then, $A_{5}$ responds the request and is willing to offer the required capability. To assess the trustworthiness of $A_{5}$, $A_{1}$ considers three factors: $direct \; trust$, $indirect \; trust$ and $reputation$ of $A_{5}$. The direct trust from $A_{1}$ to $A_{5}$ is represented as an edge $\{A_{1}\rightarrow A_{5}\}$, which is a direct trust relationship based on their past interactions. The indirect trust from $A_{1}$ to $A_{5}$ is represented as a path $\{A_{1}\rightarrow A_{2}\rightarrow A_{6}\rightarrow A_{5}\}$, which is a propagated trustworthiness based on neighbours with direct trust relationship of $A_{5}$ on the same experiences on $TK_{3}$. For example, in a crowdsourcing environment, if a task requester and a task responder has little interactions before, the requester can seek for advices from his/her previous partners to evaluate the trustworthiness of the responder on certain kinds of tasks. Then, these partners also seek for advices from their previous partners until find some one who has direct trust relationship with the requester. Reputation of $A_{5}$ in the graph is represented as a trust based on the performance of $A_{5}$ for all task categories from both neighbours and others in the environment. Therefore, the direct trust from $A_{3}$ and $A_{4}$ to $A_{5}$ should also be considered, which are represented as two edges $\{A_{3}\rightarrow A_{5}\}$ and $\{A_{4}\rightarrow A_{5}\}$, respectively.

\begin{figure}
\centering
\includegraphics[height = 0.39\textwidth, width=0.42\linewidth]{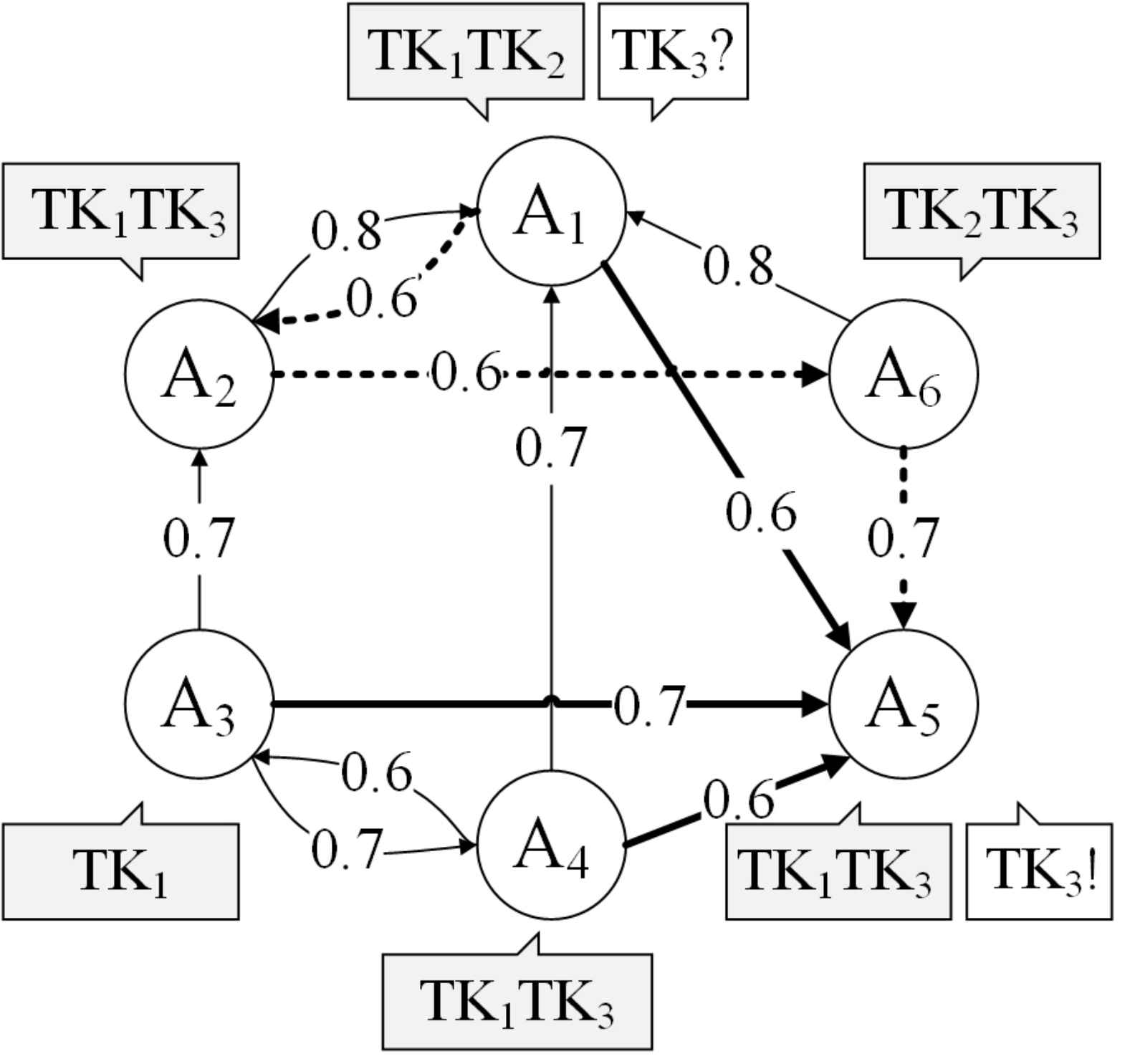}
\caption{A sample of the sociable environment.}
\label{ada_nhb}
\end{figure}

\subsection{Definitions}
The formal definitions of an environment, a task category, a task, an agent, a trustor, a trustee and an interaction are given as below.

\vspace{0.25cm}
\noindent
\textbf{Definition 1 (Environment).}  An \textit{environment} is defined as a weighted directed graph, $G = (V, E) $, where \emph{V} = \{$v_{1}, ..., v_{n}\}$ represents a set of agents; \emph{E}= \{$e_{i,j} \mid 1\leq{i}, j\leq{n}\}$ represents the trust relationship from $v_{i}$ to $v_{j}$ based on their direct interactions; the weight $w_{i,j}$ on edge $e_{i,j} \in [0, 1]$ represents trust value, where 0 and 1 represent completely untrustworthy and trustworthy, respectively. 
\vspace{0.25cm}

\noindent
\textbf{Definition 2 (Task Category).} \textit{Task Category} is defined by a set, $C = \{c_{1}, c_{1}, ..., c_{m}\}$, which represents a set of task categories.
\vspace{0.25cm} 

\noindent 
\textbf{Definition 3 (Task).} A \textit{Task} is defined as a two-tuple, $TK = (c, t)$, where $c \in C$ represents a task category; and $t \in T $ represents the required accomplish time of the task, where $T$ is a set where each element is a time point, which can be hours, or other time units depending on applications.
\vspace{0.25cm} 

\noindent
\textbf{Definition 4 (Agent).} An \textit{Agent} is defined as a three-tuple, $A = (v_{i}, C_{f}, C_{a})$, where $v_{i}$ represents an agent; $C_{f} \subset C$ represents multiple categories of tasks that $v_{i}$ previous completed, and $C_{a} \subset C$ represents multiple categories of tasks that $v_{i}$ has ability to perform.
\vspace{0.25cm}

\noindent
\textbf{Definition 5 (Trustor).} A \emph{Trustor} is defined as an agent $v_{i}$, which evaluates the trustworthiness of other agents in a sociable environment.
\vspace{0.25cm}

\noindent
\textbf{Definition 6 (Trustee).} A \emph{Trustee} is defined as an agent $v_{j}$, whose trustworthiness is evaluated by other agents in a sociable environment.
\vspace{0.25cm}

\noindent
\textbf{Definition 7 (Interaction).} An \emph{Interaction} is defined as a five-tuple, $I= (v_{i}, v_{j}, r, c, t)$, where $v_{i}, v_{j}\in{V}$ represent a trustor and a trustee, respectively; $\emph{r} \in [0,1]$ represents trustor's rating on trustee for the task interaction; \emph{c} represents a certain category of tasks; $t \in T $ represents the accomplish time of the task. 
\vspace{0.25cm}

\section{Principle and Detail Design of the Robust Trust Model}
\subsection{Overview of the Trust Model}
Our trust model is composed of three modules: Direct Trust (DT), Indirect Trust (IT) and Reputation (RT), shown in Fig. 2. 
\vspace{0.25cm}

\begin{figure}
\centering
\includegraphics[height = 0.17\textwidth, width=0.5\linewidth]{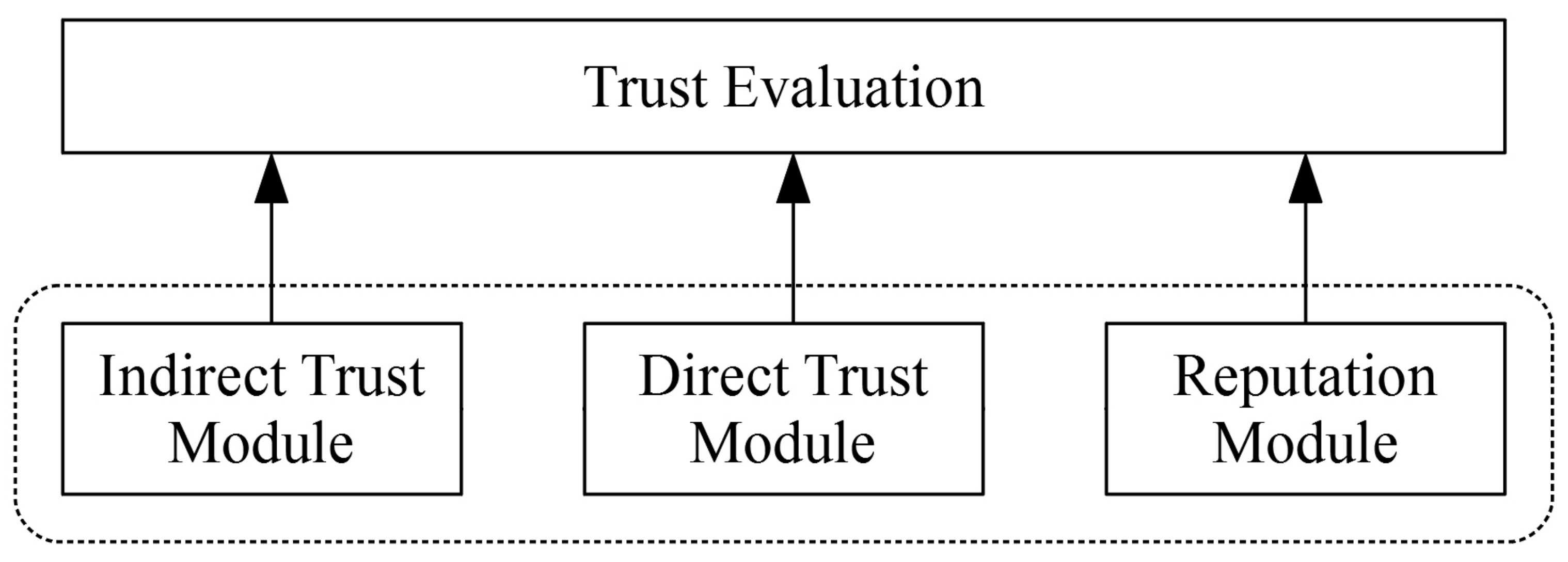}
\caption{Structure of the trust model.}
\label{ada_nhb}
\end{figure}

\noindent
\textbf{DT Module:} Direct Trust evaluation is the most widely used method in a sociable environment. The DT module will evaluate the trust value based on past direct interactions with a potential partner for the requested task.  
\vspace{0.25cm}

\noindent
\textbf{IT Module:} Indirect Trust can strengthen the evaluation by taking advantage of recommendations of other agents, who have direct interactions with the potential partner for the requested tasks. The IT module will evaluate the indirect trust value through the propagation of related agents from the network.
\vspace{0.25cm}

\noindent
\textbf{RT Module:} Reputation can strengthen the evaluation by using the knowledge from other members who had interactions with a trustee in all tasks. It is also very useful to evaluate newcomers, who have no interaction with others. The RT module will produce a reputation value for a member in the environment.
\vspace{0.25cm}

A weighted average method is used to combine the three modules, which is formally defined by Equation 1. 
\begin{equation}
Trust = \alpha T_{DT} + \beta T_{IT} + (1 - \alpha - \beta)T_{RT}
\end{equation}

The calculation of $\alpha$ and $\beta$ are formally defined by Equation 2 and Equation 3, respectively.

\begin{equation}
\alpha = 
\left\{
\begin{array}{lr}
\frac{N^{{c}'}_{DT}}{2 \times N^{c}_{DT_{min}}}, & N^{c}_{DT} = 0 \wedge N^{{c}'}_{DT} < N^{c}_{DT_{min}}, {c}' \in C_{f} \backslash c\\
\frac{1}{2}, & N^{c}_{DT} = 0 \wedge N^{{c}'}_{DT} \geq N^{c}_{DT_{min}}, {c}' \in C_{f} \backslash c\\
\frac{N^{c}_{DT}}{N^{c}_{DT_{min}}}, & 0 < N^{c}_{DT} < N^{c}_{DT_{min}}\\ 
1, & N^{c}_{DT} \geq N^{c}_{DT_{min}}
\end{array}
\right.
\end{equation}

\begin{equation}
\beta = 
\left\{
\begin{array}{lr}
0, & c \notin C_{f} \wedge c \in C_{a}\\
(1 - \alpha) \times \frac{N^{c}_{IT}}{N^{c}_{DT_{min}}}, & N^{c}_{IT} < {N^{c}_{DT_{min}}}\\ 
1 - \alpha, & N^{c}_{IT} \geq {N^{c}_{DT_{min}}}
\end{array}
\right.
\end{equation}
\vspace{0.25cm} 

where $N^{c}_{DT}$ and $N^{{c}'}_{DT}$ represents the interaction number between a trustor and a trustee on task category $c$ and other categories ${c}'$; $N^{c}_{DT_{min}}$ represents the minimum interaction number required for trust evaluation based only on direct trust, which is assigned as the average interaction number of all the memembers on $c$. $N^{c}_{IT}$ represents the number of indirect trust path between a trustor and a trustee on $c$. The minimum path number required for trust evaluation based on indirect trust is also assigned as $N^{c}_{DT_{min}}$. For example, if a trustee is a newcomer, then $N^{c}_{DT} = N^{{c}'}_{DT} = 0$, and $c \notin C_{f} \wedge c \in C_{a}$. Therefore, $\alpha = \beta = 0$, and trustee's trustworthiness is assigned as the average reputation of members in the environment. Detailed designs of three modules are introduced by the following three subsections, respectively.

\subsection{Detail Design of Direct Trust Module}
Direct Trust comes from the direct interactions between a trustor and a trustee. To evaluate the trustworthiness of a trustee on certain category of tasks $c$ in time $t$, a trustor's previous interactions with the trustee on category $c$ before time $t$ should be selected. If there is no interaction on category $c$, then interactions on other task categories ${c}'$ should be considered, because it partially influences the trustworthiness of a trustee on category $c$. The indirect trust calculation process is formally represented by Algorithm 1:

\vspace{0.25cm}
\begin{algorithm}[H]
\caption{Direct Trust Calculation}\label{algorithm}
\KwData{Interaction $I$, task category $c$, trustor $tr$, trustee $te$, time $\theta_{t}$}
\KwResult{Indirect Trust $IT$}
\uIf{$length(I_{tr, te, c, t < \theta_{t}}(:, 2)) > 0$}{
	$T_{DT} = \frac{\sum_{t < \theta_{t}}I_{tr, te, c, t}(:, 2)\times d_{t}} {\sum_{t < \theta_{t}} d_{t}}$\;
}
\ElseIf{$length(I_{tr, te, t < \theta_{t}}(:, 2)) > 0$}{
	\For{${c}' \in C \backslash c$}{ 
	$T_{DT_{c}'} = \frac{\sum_{t < \theta_{t}}I_{tr, te, {c}', t}(:, 2)\times d_{t}} {\sum_{t < \theta_{t}} d_{t}}$\;
	}
	$T_{DT} = \frac{\sum_{{c}' \in C \backslash c}T_{DT_{{c}'}}} {count({c}')}$\;
}
\end{algorithm}
\vspace{0.25cm}

The Algorithm 1 is explained as follows.

\noindent
\textbf{Step 1: (Line 1-2)} When a trustor $tr$ calculates the trustworthiness of a trustee $te$ on task category $c$ in time $\theta_{t}$, interaction $I$ is checked to find if $tr$ has interactions with $te$ on category $c$ before time $\theta_{t}$. If $tr$ and $te$ have interactions before, the direct trust $T_{DT}$ is calculated based on past interactions and discounting factor $d_{t}$, which represents $tr$'s rating on $te$ decays with time.

\noindent
\textbf{Step 2: (Line 3-8)} If $tr$ and $te$ have no interaction on task category $c$ before $\theta_{t}$, the direct trust on other categories ${c}'$ is calculated based on $I$ and $d_{t}$. Then, the average value of the direct trust on other categories ${c}'$ is assigned to $T_{DT}$.

\subsection{Detail Design of Indirect Trust Module}
Indirect trust comes from trust propagation between a trustor and a trustee through their neighbours. However, there exist some difficulties in indirect trust evaluation. As shown in Fig. 3, firstly, a trustor might have many neighbours but only a few of them have direct or indirect trust relationship with the trustee. Secondly, there might exist multiple trust propagation paths between a trustor and a trustee. Thirdly, the trust evaluation time is usually limited in real life. To solve these problems, two mechanisms are proposed as follows. 

\vspace{0.25cm}
\noindent
\textbf{Indirect Trust Propagation.} Two concepts are proposed in this module, \emph{trusted  neighbour} and \emph{propagation probability}. Besides, a tree-based algorithm is proposed to search for multiple trust propagation paths from a trustor to a trustee in a limited time period.
\vspace{0.25cm}

\noindent
\textbf{Indirect Trust Aggregation.} Based on the above trust propagation paths, an aggregation algorithm is proposed to combine all these paths and calculate the indirect trust between a trustor and a trustee.
\vspace{0.25cm}

\begin{figure}
\centering
\includegraphics[height = 0.23\textwidth, width=0.53\linewidth]{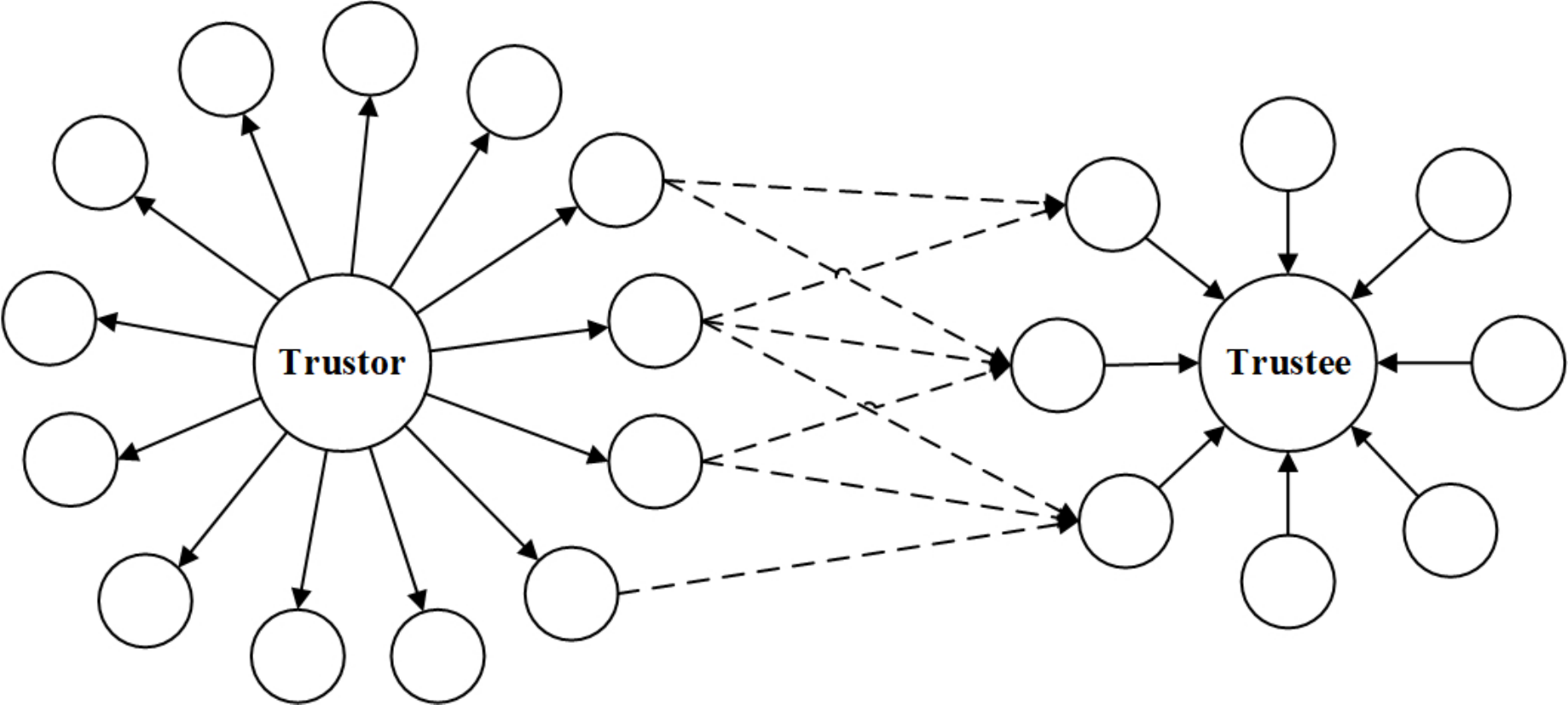}
\caption{A sample of indirect trust propagation.}
\label{ada_nhb}
\end{figure}

A trusted neighbour denotes an agent that has experiences on certain category of tasks and is highly trusted by a trustor, which can be selected from trustor's neighbours with high trustworthiness. For example, in a movie recommendation network, if a user wants to know the quality of a movie, he/she tends to seek for advices from friends who like to watch such kind of movies because they are more likely to have seen this movie or have some understanding of this movie from their friends' recommendations.

\vspace{0.25cm}
\noindent
\textbf{Definition 8 (Trusted Neighbour).} The \textit{trusted neighbour} $N_{i}^{c}= \{v_{j} \mid 1\leq{j}\leq{n}\}$ represents agent $v_{i}$'s trustworthy neighbour $v_{j}$ on task category $c$, where the weight $w_{i,j}$ on edge $e_{i,j} \in G(E)$ is higher than a given trust threshold $\theta_{r}$. The value of weight $w_{i,j}$ is the average of direct trust from $v_{i}$ to $v_{j}$ on all task categories.
\vspace{0.25cm}

Propagation probability denotes the probability of a trustor to seek advices from its trusted neighbours on certain category of tasks, which is influenced by neighbour's interaction frequency on certain task category. For example, in an e-commerce environment, if a buyer wants to know the trustworthiness of a seller in certain kinds of products, he/she tends to seek for advices from friends who purchased these products frequently in recent time, because they are more likely to have traded with this seller.

\vspace{0.25cm}
\noindent
\textbf{Definition 9 (Propagation Probability).} The \textit{propagation probability} $P_{i}^{c} = \{p_{i,j}^{c}\mid 1\leq{j}\leq{n}\}$ represents the probability of agent $v_{i}$ to seek for advices from its trusted neighbour $v_{j}$ on task category $c$. Formal definition is defined as follows, where $p^{c}_{n}$ and $p^{c}_{t}$ are two probability factors which can be calculated based on the interaction number and interaction time of $v_{j}$ on task category $c$ in specific sociable environments. For example, for high frequency interaction environments such as e-commerce and social network, nonlinear functions like logarithmic function can be used to normalize the probability value into [0, 1]. 
\vspace{0.25cm}

\begin{equation}
P_{i}^{c} = 
\left\{
	\begin{array}{lr}
	p_{i,j}^{c} = p^{c}_{n} \times p^{c}_{t}, &  0\leq{p^{c}_{n}}, p^{c}_{t}\leq{1}\\
	\sum_{j} p_{i,j}^{c} =1, &  c \in C, v_{j} \subseteq N_{i}^{c}
	\end{array}
\right.
\end{equation}

\begin{figure}
	\centering
	\includegraphics[height = 0.2\textwidth, width=0.55\linewidth]{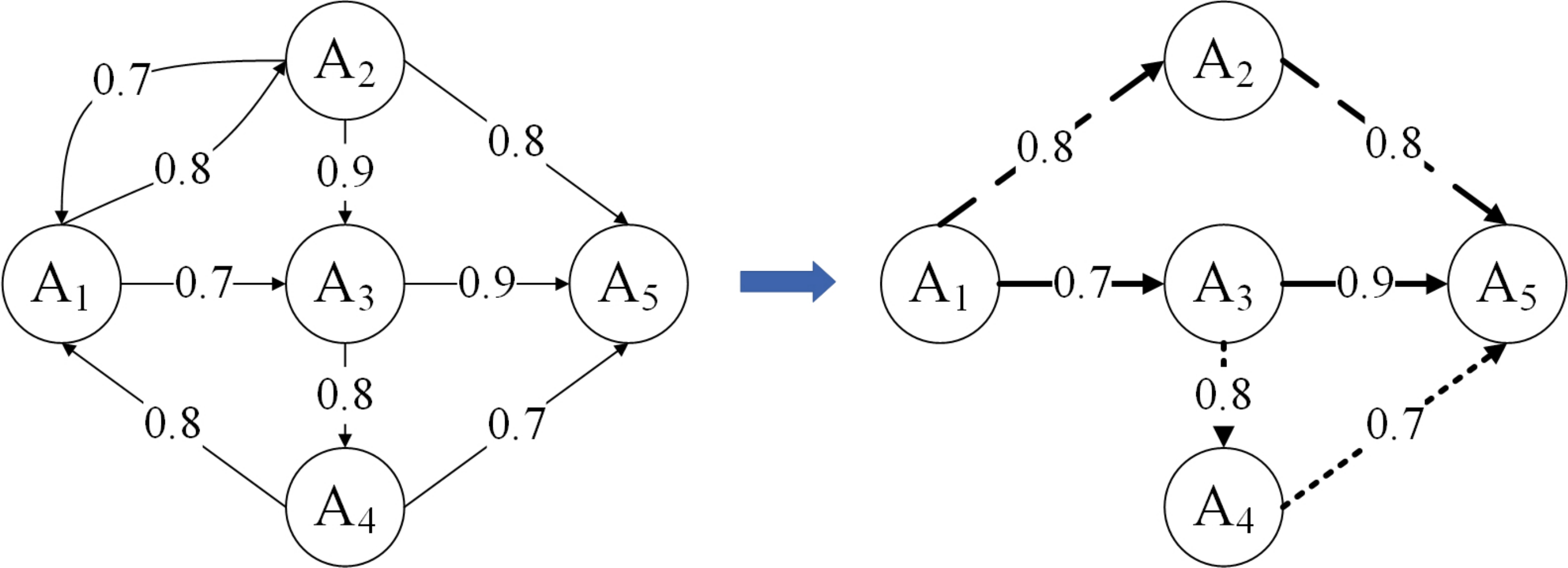}
	\caption{A sample of two principles in indirect trust propagation.}
	\label{ada_nhb}
\end{figure}

The following two principles are used in indirect trust propagation process:

\noindent
\textbf{(1) There is no loop in the trust path.} 

The trust path should start from a trustor and end with a trustee. There are two loops in Fig. 4, i.e. $\{A_{1}\rightarrow A_{2}\rightarrow A_{1}\}$ and $\{A_{1}\rightarrow A_{3}\rightarrow A_{4}\rightarrow A_{1}\}$. In this situation, the loop could be disconnected by deleting edges $e_{2,1}$ and $e_{4,1}$. 
\vspace{0.25cm}

\noindent
\textbf{(2) Direct trust path is more reliable than indirect trust path.} 

An agent tends to trust its own interactions than recommendations. In Fig. 4, there are two paths between  $A_{1}$ and $A_{3}$, i.e. $\{A_{1}\rightarrow A_{3}\}$ and $\{A_{1}\rightarrow A_{2}\rightarrow A_{3}\}$. In this situation, the indirect path should be neglected by deleting edge $e_{2,3}$. 

By combining trusted neighbours, propagation probability and two principles, a tree-based searching algorithm is proposed to find multiple trust propagation paths between a trustor and a trustee. In this module, time threshold $\theta_{t_{TK}}$ is used to avoid too long path from a trustor to a trustee, which is assigned based on the required accomplish time of certain task. Fig. 5 shows a sample of the searching process, which is formally defined by Algorithm 2.

\begin{figure}
\centering
\includegraphics[height = 0.28\textwidth, width=0.86\linewidth]{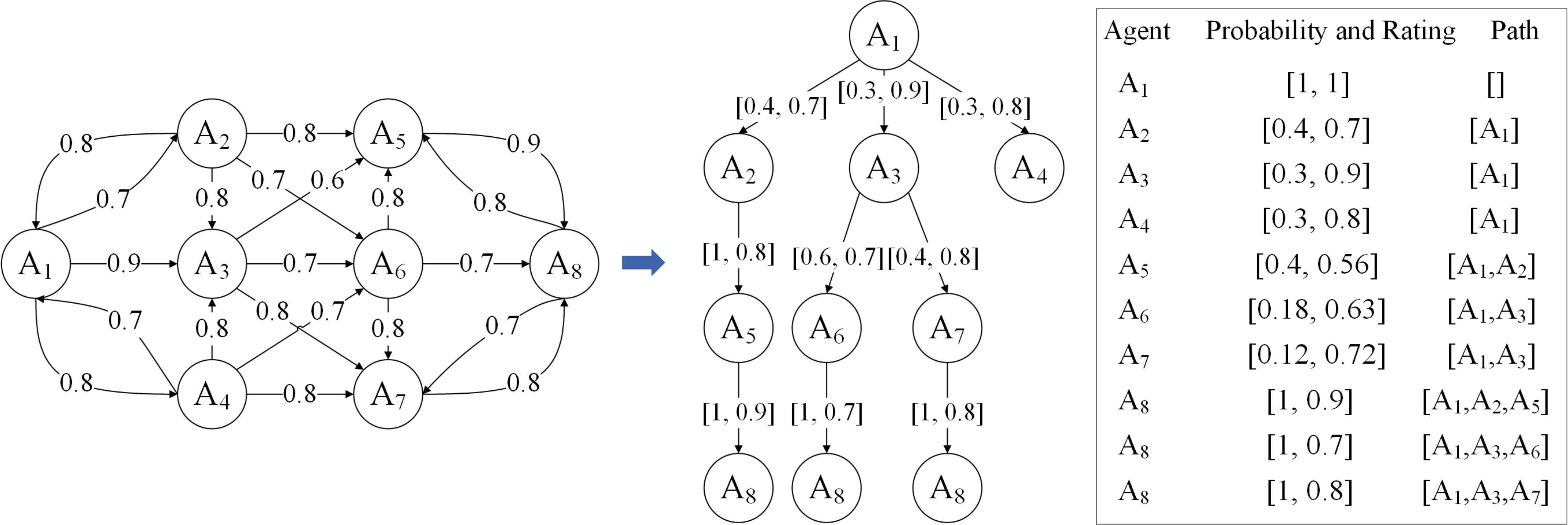}
\caption{A sample of searching for multiple trust propagation paths.}
\label{ada_nhb}
\end{figure}

\begin{algorithm}[H]
\caption{Indirect Trust Propagation}\label{algorithm}
\KwData{Environment $G$, agent $A$, interaction $I$, task category $c$, trustor $tr$, trustee $te$, trust threshold $\theta_{r}$, time $\theta_{t}$, time threshold $\theta_{t_{TK}}$}
\KwResult{Local indirect trust table $Table$}
Table = $[tr, [1,1], []]$; enqueue (Q, $tr$)\;
\While{$Q \neq \varnothing \wedge$ getTime() $ < \theta_{t_{TK}}$}{
	Sort $Q$ according to $Table(:,1)[:,0] \times Table(:,1)[:,1]$ from high to low\;
	$v_{i}$ = dequeue $Q$\;
	\For{each neighbour $v_{j}$ of $v_{i}$}{
		\uIf{$v_{j} \neq te$}{
			\uIf{$Table_{v_{j}} == \varnothing \wedge c \in A_{v_{j}}(1) \wedge w_{i,j}\geq \theta_{r}$}{
				add($N_{i}^{c}$, $v_{j}$)\;	
			}
			\ElseIf{$Table_{v_{j}} \neq \varnothing \wedge c \in A_{v_{j}}(1) \wedge w_{i,j}\geq \theta_{r}$}{
				\If{$Table_{v_{j}}(:,2) \not\subset Table_{v_{i}}(:,2) \wedge Table_{v_{j}}(:,1)[:,1] < Table_{v_{i}}(:,1)[:,1] \times w_{i,j}$}{
					$Table_{v_{k}}(:,1)[:, 0] = Table_{v_{k}}(:,1)[:, 0] \times \frac{1}{1 - Table_{v_{j}}(:,1)[:, 0]}$ subject to $Table_{v_{j}}(:,2) \subset Table_{v_{k}}(:,2)$\;
					add($N_{i}^{c}$, $v_{j}$); delete($Table_{v_{j}})$\;
				}
			}
		}
		\Else{
		rating = $\overline{I_{v_{i}, te, c, t < \theta_{t}}(:, 2)}$\;
		add(Table, $[te, [1, rating], [Table_{v_{i}}(2), v_{i}]$)\;
		}
	}
	\For{each trusted neighbour $v_{m}$ in $N_{i}^{c}$}{
		calculate $p_{i,m}^{c}$; enqueue $(Q, v_{m})$\;
		add(Table, $[v_{m}, Table_{v_{i}}(1) \times [p_{i,m}^{c}, w_{i,m}], [Table_{v_{i}}(2), v_{i}]]$)\;
	}			
}
\end{algorithm}
\vspace{0.25cm}

The Algorithm 2 is explained as follows.

\noindent
\textbf{Step 1: (Line 1)} When a trustor $tr$ calculates the trustworthiness of a trustee $te$, $Table$, $Q$ is created to store the information of propagation paths and agents on the paths, respectively. Specifically, $Table$ is composed of three columns: agents $v_{i}$ on the path, cumulative propagation probability and cumulative trust, and path from $tr$ to $v_{i}$. If $v_{i}$ is $te$, then the propagation probability and cumulative trust are set as 1 and current trust rating, respectively.

\noindent
\textbf{Step 2: (Line 2-4)} When $Q$ is not null and current time is less than threshold $\theta_{t_{TK}}$, $Q$ is sorted according to the product of cumulative propagation probability and cumulative trust in $Table$, where $\theta_{t_{TK}}$ is assigned based on the required task accomplish time. Agent $v_{i}$ with the highest product value is taken out of $Q$.

\noindent
\textbf{Step 3: (Line 5-14)} For $v_{i}'s $ each neighbour $v_{j}$, if it is not $te$ and meets three conditions: (1) not in $Table$, which means it has not been visited; (2) having interactions in task category $c$; and (3) highly trusted by $v_{i}$, then $v_{j}$ is added into $v_{i}$'s trusted neighbours $N_{i}^{c}$. If $v_{j}$ is in $Table$, meets conditions (2) and (3), and meets the other three conditions: i.e. (4) there is no loop between $v_{i}$ and $v_{j}$; (5) there is no direct path between $v_{j}$ and $v_{i}$'s ancestors; and (6) $v_{j}'s $ current cumulative trust is higher than that of old one in $Table$, then the cumulative propagation probability of $v_{j}$'s siblings and siblings' offsprings $v_{k}$ is revised. Besides, $v_{j}$ is added into $N_{i}^{c}$, and the record of $v_{j}$ is deleted from the $Table$.  

\noindent
\textbf{Step 4: (Line 15-19)} If $v_{j}$ is $te$, the trust rating from $v_{i}$ to $te$ before time $\theta_{t}$ could be calculated. Then, a record containing $te$, rating and path between $tr$ to $te$ is added into the $Table$.

\noindent
\textbf{Step 5: (Line 20-24)} For each trusted neighbour $v_{m}$ in $N_{i}^{c}$, the propagation probability from $v_{i}$ to $v_{m}$ is calculated based on $v_{m}$'s interactions on category $c$. Then, $v_{m}$ is added into $Q$. A record containing $v_{m}$, the cumulative propagation probability and cumulative trust, and path between $tr$ to $v_{m}$ is added into the $Table$. Going to Step 2.

An aggregation algorithm is proposed to aggregate multiple indirect trust paths between a trustor and a trustee. In this algorithm, a propagation path connecting a trustor and a trustee is divided into two parts: $\{trustor \rightarrow ... \rightarrow A_{i}\}$ and $\{A_{i} \rightarrow trustee\}$. The cumulative trust on each part represents path trustworthiness and rating on $trustee$, respectively. For example, in Fig. 5, if the trust threshold $\theta_{r_{p}} = 0.6$, then the indirect trust from $A_{1}$ to $A_{8} = \frac{0.7 \times 0.63 + 0.8 \times 0.72}{0.63 + 0.72} = 0.75$. A formal aggregation process is presented by Algorithm 3. 

\begin{algorithm}[H]
\caption{Indirect Trust Aggregation}\label{algorithm}
\KwData{Local indirect trust table $Table$, trustee $te$, trust threshold $\theta_{r_{p}}$, decay factor $d$}
\KwResult{Indirect trust $IT$}
\If{$Table_{te}\neq \varnothing$}{
	\For{each path $i$ in $Table_{te}$}{
		$n_{i} = Table_{te}(i,2)[-1]$;
		$r_{i} = Table_{te}(i,1)[:,1]$;
		add$(IT_{N}, [n_{i}, r_{i}]$)\;
	}
	\For {each neighbour $v_{i}$ in $IT_{N}(0)$}{
		$w_{tr,v_{i}} = Table_{v_{i}}(1)[:,1]$\;
		add$(IT_{Path}, [IT_{N_{v_{i}}}(1), w_{tr,v_{i}}])$ subject to $w_{tr,v_{i}}> \theta_{r_{p}}$\;
	}
	$path_{len} = length(IT_{Path})$\; 
	IT = $\frac{\sum_{0 \leq i < path_{len}} IT_{Path}(i,0) \times IT_{Path}(i,1) } {\sum_{0 \leq i < path_{len}} IT_{Path}(i,1) }$ subject to $path_{len} > 1$\;
	$len = length(Table_{IT_{Path}(0)}(2)) + 1$;
	$IT = IT_{Path}(1) \times d^{len}$ subject to $path_{len} == 1$\;
}
\end{algorithm}
\vspace{0.25cm}

The Algorithm 3 is explained as follows.

\noindent
\textbf{Step 1: (Line 1-4)} If trustee $te$ is in $Table$, which means there has paths between $tr$ and $te$. For each path $i$ in $Table_{te}$, neighbour $n_{i}$, trust rating $r_{i}$ from $n_{i}$ to $te$ are added into $IT_{N}$. 

\noindent
\textbf{Step 2: (Line 5-8)} For each neighbour $v_{i}$ in $IT_{N}$, $v_{i}$'s cumulative trust is selected from $Table$. If the cumulative trust is higher than the threshold $\theta_{r_{p}}$, it means that the propagation path is trustworthy, then $v_{i}$ and its cumulative trust value are added into the $IT_{Path}$.  

\noindent
\textbf{Step 3: (Line 9-12)} If there are multiple paths in $IT_{Path}$, the indirect trust $IT$ is calculated based on $IT_{Path}$. If there is only one path, the value of indirect trust $IT$ is calculated based on the decay factor $d$ and $IT_{Path}$.

\subsection{Detail Design of Reputation Module}
Reputation comes from members who have interactions with the trustee for all tasks. However, it is difficult to aggregate trust evaluation from different agents with unknown overall trustworthiness. To solve this problem, the trust evaluation is viewed as a reputation propagation process from a trustor to a trustee. However, there are two difficulties in the reputation evalulation. Firstly, a trustor should not propagate its reputation to untrusted trustees. Secondly, a trustor should propagate more reputation to trustees with higher trustworthiness. Therefore, a trust threshold $\theta_{r}$ is used to select trusted trustees, and a reputation propagation ratio is calculated before using the PageRank method. A formal reputation calculation process is presented by Algorithm 4. 

\vspace{0.25cm}
\begin{algorithm}[H]
	\caption{PageRank-based Reputation Calculation}\label{algorithm}
	\KwData{Environment $G$, Interaction $I$, trustee $te$, trust threshold $\theta_{r}$, time threshold $\theta_{t}$, damping factor $q$, convergence factor $\epsilon$}
	\KwResult{Reputation $RT$}
	\For{$each \; v_{i} \; in \; G(V)$}{
		$add(V_{R}, v_{i})$ subject to $w_{j, i} >= \theta_{r}: v_{j} \in G(V) \backslash v_{i}, \; w_{j, i} \in G(E)$\;
	}
	\For{$each \; v_{i} \; in \; V_{R}$}{
		$r_{max} = max(w_{i,j}) : v_{j} \in V_{R} \backslash v_{i}, w_{i,j} \in G(E)$\;
		\For{ each neighbour $v_{j}$ of $v_{i}$}{
			${w}'_{i,j} = \frac{w_{i,j} \times r_{max}} {\sum{w_{i,j}}}$ subject to $w_{i,j} >= \theta_{r}$\;
			${w}'_{i,j} = \frac{1 - r_{max}} {count(w_{i,j})}$ subject to $w_{i,j} < \theta_{r}$\;
			add$(E_{R}, {w}'_{i,j})$\;
		}
	}
	e = [$\frac{1}{length(V_{R})}$, ..., $\frac{1}{length(V_{R})}]_{1 \times length(V_{R})};$ $t = getTime()$\;
	$R$ = $q \times E_{R}^{T} \times e^{T} + (1-q) \times e^{T}; \; {R}'$ = $q \times E_{R}^{T} \times R + (1-q) \times e^{T}$\;
	\While{$\left | {R}' - R \right | > \epsilon \; \wedge  t < \theta_{t}$}{
		$R = {R}'; \; {R}''$ = $q \times E_{R}^{T} \times {R}' + (1-q) \times e^{T}$\;
		${R}' = {R}''; \; t = getTime()$\;
	}
	normalize ${R}'$ to [0, 1];
	$RT =  {R}'(te)$ subject to $te \in V_{R}$\;
	$RT = average({R}')$ subject to $te \notin V_{R}$\;
\end{algorithm}
\vspace{0.25cm}

The Algorithm 4 is explained as follows.

\noindent
\textbf{Step 1: (Line 1-3)} For each agent $v_{i}$ in $G(V)$, if $w_{j, i} > \theta_{r}$, which means that $v_{i}$ has been evaluated with a rating higher than the threshold, then $v_{i}$ is added to node set $V_{R}$.

\noindent
\textbf{Step 2: (Line 4-11)} For each $v_{i}$ in $V_{R}$, the maximum rating $r_{max}$ is selected as a ratio that $v_{i}$ propagate its reputation to trustworthy neighbour $v_{j}$ ($w_{i,j} > \theta_{r}$). Then, $v_{i}$'s remaining reputation $1 - r_{max}$ is propagated to others in $V_{R}$ equally. After that, ${w}'_{i,j}$ is added into edge set $E_{R}$.

\noindent
\textbf{Step 3: (Line 12-17)} With $V_{R}$ and $E_{R}$, a PageRank based reputation model can be created. Specifically, the initial reputation vector $e$ of each agent in $V_{R}$ is assigned to $\frac{1}{length(V_{R})}$, the damping factor $q$ is usually assigned to 0.85 in PageRank algorithm. Based on $e$, $q$ and the reputation propagation matrix $E^{T}_{R}$, a reputation vector $R$ can be calculated. Then, a new reputation vector ${R}'$ is calculated based on the previous one $R$. Iteratively calculating new reputation vector ${R}'$ untill convergence or time $t$ is higher than threshold $\theta_{t}$. 
	
\noindent
\textbf{Step 4: (Line 18-19)} ${R}'$ is normalized to [0, 1]. If trustee $te \in V_{R}$, its reputation value in ${R}'$ returns. Otherwise, an average value of ${R}'$ returns.

\section{Related Work}

A lot of trust evaluation methods and models have been developed. For example, SocialMF proposed by Jamali and Esteris \cite{Jamali2010socialmf} is a trust-aware algorithm which has been widely used in recommendation systems, with the consideration of  trust propagation in the matrix factorization approach, and calculated target users' feature vector based on the features of its direct neighbours in social networks. However, SocialMF does not consider the impact of untrusted neighbours, which widely exist in sociable environment. In this paper, a trust threshold is used to select trustworthy neighbours and prevent the impact of untrusted neighbours in trust propagation process. VectorTrust proposed by Zhao and Li \cite{zhao2013aggregation} is a local trust management scheme for P2P networks, which used a trust vector to represent trust, and propagated indirect trust through trust paths. However, VectorTrust only considers the most trustworthy path when facing multiple paths, and neglects the impact of trust subjectivity. In this paper, a tree-based searching algorithm and an aggregation algorithm are used to find and combine multiple trustworthy paths together, so as to reduce the impact of trust subjectivity. Yan and et al \cite{yan2015graph} proposed a graph-based reputation model, which used some social context, including users' activities and relationships to calculate trust in case of limited first-hand information like a newcomer. However, their model is designed for a specific domain, i.e. social commerce, which might not suit for other sociable environments where agents interact with each other frequently.

\section{Conclusion}
In this paper, a new model for trust evaluation between a trustor and a trustee in a sociable environment was proposed. The proposed model is robust since the model evaluates a trust value for a trustee by utilization of direct trust, indirect trust and reputation. The model design also considered multiple task categories and decentralized manner, so our model can be used in different types of social networks and applications. In the future, we will firstly test our model by different data sets to evaluate the performance in terms of impacts from direct, indirect and reputation to the trust evaluation. Secondly, we will test our model in the real-world data sets for its applications.

\section*{Acknowledgments}
This research is supported by the UPA and IPA scholarships from the University of Wollongong for the Joint PhD Program with Nagoya Institute of Technology.
\bibliographystyle{splncs03}
\bibliography{bib}

\end{document}